\documentclass{ieeeaccess}
\usepackage{amsmath,amssymb,amsfonts}
\usepackage{algorithm, algorithmicx, algpseudocode}
\usepackage{amsmath}
\usepackage{graphicx}
\usepackage{textcomp}

\usepackage{bm}
\makeatletter
\AtBeginDocument{\DeclareMathVersion{bold}
\SetSymbolFont{operators}{bold}{T1}{times}{b}{n}
\SetSymbolFont{NewLetters}{bold}{T1}{times}{b}{it}
\SetMathAlphabet{\mathrm}{bold}{T1}{times}{b}{n}
\SetMathAlphabet{\mathit}{bold}{T1}{times}{b}{it}
\SetMathAlphabet{\mathbf}{bold}{T1}{times}{b}{n}
\SetMathAlphabet{\mathtt}{bold}{OT1}{pcr}{b}{n}
\SetSymbolFont{symbols}{bold}{OMS}{cmsy}{b}{n}
\renewcommand\boldmath{\@nomath\boldmath\mathversion{bold}}}

\makeatother

\def\BibTeX{{\rm B\kern-.05em{\sc i\kern-.025em b}\kern-.08em
    T\kern-.1667em\lower.7ex\hbox{E}\kern-.125emX}}

\begin{document}
\history{Date of publication xxxx 00, 0000, date of current version June 23, 2025.}
\doi{10.1109/ACCESS.2023.1120000}

% \author{Ruhaan Singh, Hong Jun Yoon, Sreelekha Guggilam}

\title{Iterative Misclassification Error Training (IMET): An Optimized Neural Network Training Technique for Image Classification}
\author{\uppercase{Ruhaan Singh}\authorrefmark{1},
\uppercase{Sreelekha Guggilam}\authorrefmark{2}}

\address[1]{Farragut High School, Knoxville, TN 37934 USA (e-mail: risruh@gmail.com)}
\address[2]{Department of Mathematics and Statistics, Texas A\&M University, Corpus Christi, TX 78412 USA (e-mail: sreelekha.guggilam@tamucc.edu)}

% \tfootnote{This paragraph of the first footnote will contain support
% information, including sponsor and financial support acknowledgment. For
% example, ``This work was supported in part by the U.S. Department of
% Commerce under Grant BS123456.''}

% \markboth
% {Author \headeretal: Preparation of Papers for IEEE TRANSACTIONS and JOURNALS}
% {Author \headeretal: Preparation of Papers for IEEE TRANSACTIONS and JOURNALS}

\corresp{Corresponding author: Sreelekha Guggilam (e-mail: sreelekha.guggilam@tamucc.edu).}

\begin{abstract}
Deep learning models have proven to be effective on medical datasets for accurate diagnostic predictions from images. However, medical datasets often contain noisy, mislabeled, or poorly generalizable images, particularly for edge cases and anomalous outcomes. Additionally, high quality datasets are often small in sample size that can result in overfitting, where models memorize noise rather than learn generalizable patterns. This in particular, could pose serious risks in medical diagnostics where the risk associated with mis-classification can impact human life.
Several data-efficient training strategies have emerged to address these constraints. In particular, coreset selection identifies compact subsets of the most representative samples, enabling training that approximates full-dataset performance while reducing computational overhead. On the other hand, curriculum learning relies on gradually increasing training difficulty and accelerating convergence. However, developing a generalizable difficulty ranking mechanism that works across diverse domains, datasets, and models while reducing the computational tasks and  remains challenging.
In this paper, we introduce Iterative Misclassification Error Training (IMET), a novel framework inspired by curriculum learning and coreset selection. The IMET approach is aimed to identify misclassified samples in order to streamline the training process, while prioritizing the model's attention to edge case senarious and rare outcomes. The paper evaluates IMET's performance on benchmark medical image classification datasets against state-of-the-art ResNet architectures. The results demonstrating IMET's potential for enhancing model robustness and accuracy in medical image analysis are also presented in the paper.
\end{abstract}

\titlepgskip=-21pt

\maketitle
\section{Introduction}
AI has been influencing healthcare by harnessing its capabilities to study vast amounts of unstructured medical data for many impactful applications, such as detection or prediction of diseases before they can reach an advanced stage, drug discovery, and treatment management. The economic implications of this is substantial. 
According to the data platform Statista, the global artificial intelligence healthcare market will be worth \$187 billion by 2030, compared to \$11 billion in 2021\footnote{https://www.ibm.com/think/insights/ai-healthcare-benefits}. This dramatic increase reflects the growing recognition of AI's potential to address critical healthcare challenges.

Given the social and economic impacts, the necessity of more informed and optimal AI-driven solutions becomes even more challenging when implementing on medical datasets. In addition, machine learning (ML) and AI offer solutions that are integral in alleviating diagnostic errors that often result in loss of life and increased health costs. Estimations by the Government Accountability Office indicate that annually 12 million Americans are misdiagnosed, resulting in around \$100 billion of associated costs\footnote{https://www.gao.gov/assets/gao-22-104629.pdf}. The human cost of this is even more concerning as found in the National Academy of Medicine report that indicate diagnostic errors contributing to approximately 10 percent of patient deaths \footnote{https://www.ncbi.nlm.nih.gov/books/NBK338586/}. These statistics underscore the need for a more robust models for medical research that maintain patient safety while minimizing risks to human lives.

Current AI approaches deliver key advantages in medical diagnostics, particularly in early detection, consistency in interpretation of diagnostic tests and scans, and enhancing medical care accessibility to patients in remote or medically under-served areas. Well-trained and robust models can detect patterns in medical images faster and earlier, compared to traditional methods, leading to quicker intervention. For example, \cite{kropp2023diabetic} have shown that diabetic retinopathy, the leading cause of new blindness in adults (CDC 2023), can be detected early using ML applications, resulting in reduction of blindness rates by up to 90\% \cite{padhy2019artificial}. Additionally, ML/DL models offer consistent interpretation of medical images and reduce variations in their interpretation across medical professionals. Incorporation of ML/DL in the medical field also allows the expansion of healthcare access particularly in underserved areas to a larger population, increasing healthcare equity across diverse populations.

Researchers \cite{wang2012machine, wernick2010machine} began exploring the application of deep learning in medical imaging, including the detection of abnormalities in X-rays and CT scans in the early parts of 2010. By 2015 \cite{cruz2014automatic, cirecsan2013mitosis} \cite{liu2014early}, deep learning models, particularly CNNs, demonstrated success in image-based disease detection. For instance, CNNs were employed to identify signs of diabetic retinopathy in retinal images \cite{pratt2016convolutional, abramoff2016improved}. The ImageNet competition \cite{russakovsky2015imagenet} showcased the capabilities of deep learning in image classification, inspiring researchers to apply these techniques to medical images. 

Deep learning (DL) started to be increasingly used in radiology for tasks such as tumor detection and in pathology applications, with DL models being employed to analyze histopathological slides for cancer detection \cite{sun2018integrating}. Simultaneously, natural language processing (NLP) models were developed \cite{zeng2018natural} to analyze electronic health records (EHRs), extract relevant information, and assist in disease diagnosis and prediction. By 2018, researchers had started integrating multiple modalities of data, such as combining imaging and genomic data, to enhance disease detection accuracy \cite{bi2019effective}, and by early 2020, the use of deep learning for predicting disease risk based on a combination of clinical and genetic data began to gain popularity\cite{venugopalan2021multimodal}. DL played a crucial role during the COVID-19 pandemic \cite{alazab2020covid} as well, especially with developing diagnostic tools, predicting disease spread, and analyzing medical imaging for COVID-19-related abnormalities. Continued research focuses on refining existing deep learning models, developing more interpretable models, and addressing issues related to data privacy and bias. 

Despite these significant advancements in AI-driven research and applications, the underlying technology driving these medical AI breakthroughs faces fundamental challenges that must be addressed. The surge in artificial intelligence (AI) research, particularly deep neural networks (DNNs), has revolutionized the development and use of novel artificial neural network (ANN) architectures. This is especially true for specific tasks like prediction, classification, and modeling of complex datasets, triggering advancements across fields ranging from healthcare to finance to materials science. By using thousands or even billions of hidden nodes (neurons), these models can autonomously generate rules substantial enough for tasks like predicting cancer or driving a car. 

However, this optimal performance comes at a cost of substantial training data which is a fundamental challenge for many domains, especially healthcare and medical sciences which often suffer from limited data availability for task-specific trainings. The scarcity problem is compounded by the black-box nature of ANNs making it hard to identify the performance bottlenecks in resource constrained situations. This emphasizes the crucial role of training processes in building trustworthy and understandable models, especially in the case of healthcare and medical data where model results eventually have a significant impact on human lives.

The scarcity of medical data that is viable for ANN training has been just one of the significant challenges with widespread adoption of AI in healthcare research. However, given the high risk associated with misclassification or delays in predictions in healthcare applications, training the model in a more streamlined and optimized fashion is highly important. For AI-powered applications to be effective, they need access to large amounts of data; deep learning models often rely on immense amounts of labeled data for training. According to \cite{zhu2004class} the performance of a ML/DL model depends on two significant factors: the quality of the training data and the competence of the learning algorithm. In medical sciences, obtaining well-annotated datasets can be challenging due to issues of privacy, data sharing restrictions, and the time-consuming nature of manual annotation. As such, many healthcare organizations do not have the necessary data sets to utilize AI effectively. Likewise, some diseases have very imbalanced data, where certain classes have extremely few samples due to the rareness of that class, which can negatively impact the performance of ML/DL algorithms.

Additionally, algorithms are only as good as the data they are trained on. Some complex algorithms make it challenging for healthcare practitioners to understand how AI provides specific recommendations. Deep learning models may achieve high accuracy in classification tasks without providing insights into the underlying biological or pathological mechanisms. Bridging the gap between model predictions and a deeper understanding of diseases is thus a significant challenge in medical research. Moreover, significant computing power is required for the analysis of large and complex data sets. Training complex deep learning models can be computationally intensive and requires significant resources. Access to high-performance computing infrastructure and expertise in model development can be a barrier, particularly for smaller research institutions or healthcare providers.

To address these fundamental challenges, researchers have developed sophisticated training methodologies that move beyond traditional approaches. Traditionally, ANNs have been trained by providing a sequence of randomly sampled mini-batches from training data. However, recent innovations in training optimization have shown promise in overcoming many of the limitations discussed above.

Sampling techniques and training optimization of ML/DL models has grown in importance over the years to address some of the above mentioned challenges. The most commonly used sampling technique is Synthetic Minority Oversampling Technique (SMOTE), developed by Han et al., 2005 \cite{han2005borderline}, which oversamples the minority class of a dataset in order to fix class imbalance. Lapedriza et al., 2013 \cite{lapedriza2013all} demonstrated that in a training dataset all examples are not of equal value and that sample selection is an important aspect of training ML/DL models. More recently, complicated techniques have been developed to improve model performance, such as importance sampling \cite{katharopoulos2018not}, where the most important samples are chosen based on reduced variance of gradient estimates and learning optimal sample weights \cite{santiago2021low}, which estimates each sample’s weight in order to reduce loss and subsequently improve training.

Amongst the many advances, we focus our research on two particular approaches in order to address the data efficiency and training optimization specific challenges. Methods like coreset selection address the challenges with computational costs as well as training data quality. As opposed to the traditional approach of using all available data indiscriminately, the coreset approach relies on strategically sampled subsets that maximize learning efficiency while maintaining performance comparable to full-dataset training. This is especially valuable when dealing with noisy or redundant data common in real-world medical datasets. Although coreset selection offers a computational advantage, the static sub-sampling approach can limit the flexibility to align with the model's learning \cite{hacohen2019power}. 

Curriculum learning has shown promise in addressing such challenges. The method mimics human learning for ANN training by feeding data in order of increasing difficulty. This organized training approach has shown remarkable improvements in model convergence, generalization, and overall robustness, particularly in computer vision tasks where traditional random sampling often leads to suboptimal learning trajectories \cite{wang2021survey}.

The goal of this research is to improve the accuracy of disease detection through the development of a novel training technique that allows use of fewer training samples for faster training of deep neural networks. The training process is inspired by curriculum learning \cite{wang2021survey, soviany2022curriculum} that implements training models on easier to more complex data points. Unlike curriculum learning, the proposed work uses the misclassified observations to optimize the training of the deep neural network. Through these streamlined methods, increased accuracy on medical imagery with very little data and class imbalance is looked to be achieved. Additionally, the method highlights the significance of prioritizing the distribution of data as neural networks train in order to achieve the best possible results.

\section{Background on Curriculum Learning}

Throughout the development of machine learning, human learning has served as a fundamental source of inspiration. This is evident in the evolution of algorithms, from basic regression models designed to predict future events to sophisticated neural networks that mimic the structure and function of the human brain. Human learning thrives on a continuous cycle of improvement through experience. Unlike traditional machine learning algorithms, humans don't simply process data; we actively learn from past mistakes. For instance, consider early humans. Initially, ignorance about flammable materials or fire's dangers could lead to injuries. However, through experience and teaching, they learned to manage fire more safely. This core ability to learn from errors sets human learning apart from traditional machine learning algorithms. While these algorithms can process vast amounts of data and improve their performance over time, they typically lack the ability to actively identify and rectify their mistakes. This distinction highlights the ongoing challenge of fostering true adaptive learning within artificial intelligence. In this paper, we would like to demonstrate the power of "learning from errors" in developing more stable neural networks.

\subsection{Curriculum Learning}
Curriculum learning has become a notable strategy in training ANNs that introduces data to the model based on the difficulty of training samples. While various methods enhance neural network training, curriculum learning \cite{bengio2009curriculum, wang2021survey, soviany2022curriculum} has garnered significant interest in recent years. Researchers have harnessed its potential in a wide range of applications, including supervised learning tasks in computer vision \cite{guo2018curriculumnet, jiang2014easy}, graph neural networks \cite{zhang2024curriculum}, and healthcare prediction \cite{hu2023dual}. The primary advantages of adopting CL strategies in diverse real-world scenarios include enhanced model performance on target tasks and accelerated training processes, addressing two key objectives of modern machine learning research.

The earliest work on curriculum learning has been proposed by Bengio et al. \cite{bengio2009curriculum}, where they define the curriculum as follows.

\textbf{Definition:}
    Given a training set $D$. A curriculum is a sequence of training criteria over $T$ training steps$: C=(Q_1,\dots,Q_t,\dots,Q_T)$ where each criterion $Q_t$ is a reweighting of the target training distribution $P(z)$
    \begin{equation}
         Q_t(z)\propto W_t(z)P(z),\ \forall z\in D
    \end{equation}
    such that the following three conditions are satisfied:
    \begin{enumerate}
        \item The entropy of distributions gradually increases, $$H(Q_t) < H(Q_{t+1})$$
        \vspace{-10pt}
        \item The weight for any example increases, $$W_t(z)\leq W_{t+1}(z),\ \forall z\in D$$
        \vspace{-10pt}
        \item $$Q_T(z)=P(z)$$
        \vspace{-10pt}
    \end{enumerate}

In simple terms, the authors in the \cite{bengio2009curriculum} define curriculum as a ordered sequence of subsets of the training data where each subset is designed to have increasingly difficult or complex observations (increasing entropy) and increase in sample size. Finally the reweighting of all examples, $Q_T$, is uniform and we train on the target training set

Previous research on curriculum learning focused on training machine learning algorithms in a streamlined format where the observation/training samples are in used in the order of their "difficulty". However, quantifying difficulty in training samples is not unique to data or the application making the underlying task very challenging. Current methodologies on defining the curriculum order or ranking observations based on their difficulty is not generalizable across data, domain or task. This makes the process user-specific and difficult to scale. Additionally, the learning scheme can make is extremely vulnerable to rare patters if they are not deemed important according to the set learning. This can make the models lack robustness to outliers, as they may be misclassified as “easy” and low-priority samples leading to potential learning biases or inefficiencies. Thus, we introduce an automated learning approach that addresses some of these potential challenges in learning from rare patterns in the data. 

\subsection{Coreset Selection}
While curriculum learning offers a more systematic approach in training an ANN, the approach does not address challenges with training costs, particularly in datasets with excess or redundant training samples. On the other hand, coreset selection addresses these challenges by identifying a small, representative subset of the training dataset that preserves essential patterns for effective model learning. In a classification setup, for a given training data $\mathcal{T}={(\mathbf{x_i}, y_i)}_{i=1}^N$ consists of $N$ i.i.d. samples from an underlying data distribution, a coreset $\mathcal{S^*}$ is a derived subset $\mathcal{S} \subset \mathcal{T}$ with $|\mathcal{S}|<<|\mathcal{T}|$ such that a model $\theta^\mathcal{S}$ trained on the subset $\mathcal{S}$ and the model $\theta^\mathcal{T}$ trained on the training data $\mathcal{T}$ satisfy the following:
$$\mathcal{S^*} = argmin_{\mathcal{S} \subset \mathcal{T}:\frac{|\mathcal{S}|}{|\mathcal{T}|}\leq 1-\alpha} \mathbb{E}_{\mathbf{x}, y\sim P}[\mathcal{L(\mathbf{x}, y;\theta^\mathcal{S})}],$$ where $\alpha\in (0,1)$ is a pruning ratio and $\mathcal{L}$ is a loss function\cite{moser2025coreset}.

% Gradient-based methods select samples that maximize the diversity of gradients, ensuring that the coreset captures the full spectrum of learning signals present in the original dataset .
% Geometric approaches utilize distance metrics in feature space to select representative points that maintain the geometric structure of the data distribution.
% Uncertainty-based selection identifies samples where the current model exhibits high uncertainty, focusing training on regions of the input space that require additional learning.
% Loss-based methods prioritize samples with high training loss, similar to hard example mining techniques commonly used in computer vision applications.

Several approaches have explored the use of coresets for ANN training spanning a wide range of algorithms from geometric approaches, uncertainty-based to loss-based methods\cite{feldman2020core, moser2025coreset}. These methods consistently aim to improve computational costs, training stability, and reduce storage costs, making it particularly valuable for resource-constrained environments and large-scale applications \cite{killamsetty2021retrieve, xia2022moderate, chai2023efficient}. Despite the many advantages, coreset methods also face limitations. Static coreset selection may not adapt to the evolving needs of the model during training, potentially missing important samples that become relevant as the model learns. While active learning based approaches\cite{wei2015submodularity} offer potential solutions, they introduce more computational complexity that may offset the performance advantage. However, the quality of coreset selection heavily depends on the chosen selection criterion, which may not generalize across different datasets or tasks.

% \subsubsection{Synergistic Potential of Combined Approaches}
Recognizing the complementary strengths and limitations of both curriculum learning and coreset selection, recent research has begun exploring their integration to create more robust and efficient training methodologies. While curriculum learning determines the optimal sequence for presenting training samples, coreset selection identifies which samples should be included in the training process. This synergy addresses multiple challenges simultaneously: coreset selection ensures data quality and computational efficiency, while curriculum learning optimizes the learning progression and convergence properties. 

Thus, we introduce an automated learning approach that addresses some of these potential challenges in learning from rare patterns in the data, leveraging insights from both curriculum learning and coreset selection methodologies.

\subsection{Overview of CNN Models}
\begin{figure}
     \centering
         \centering
         \includegraphics[width=0.45\textwidth]{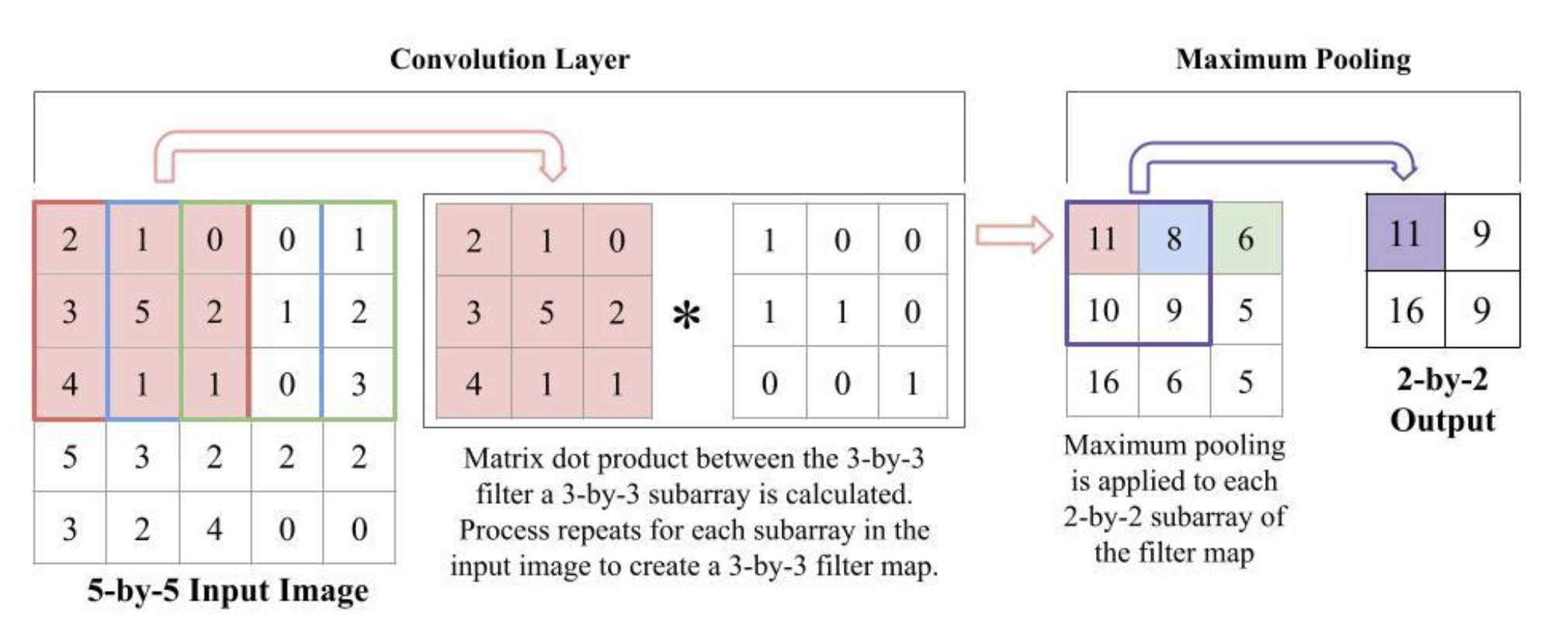}
         \caption{Example process of convolutional and pooling layers}
         \label{fig:1}
\end{figure}

\begin{figure}
     \centering
         \centering
         \includegraphics[width=0.45\textwidth]{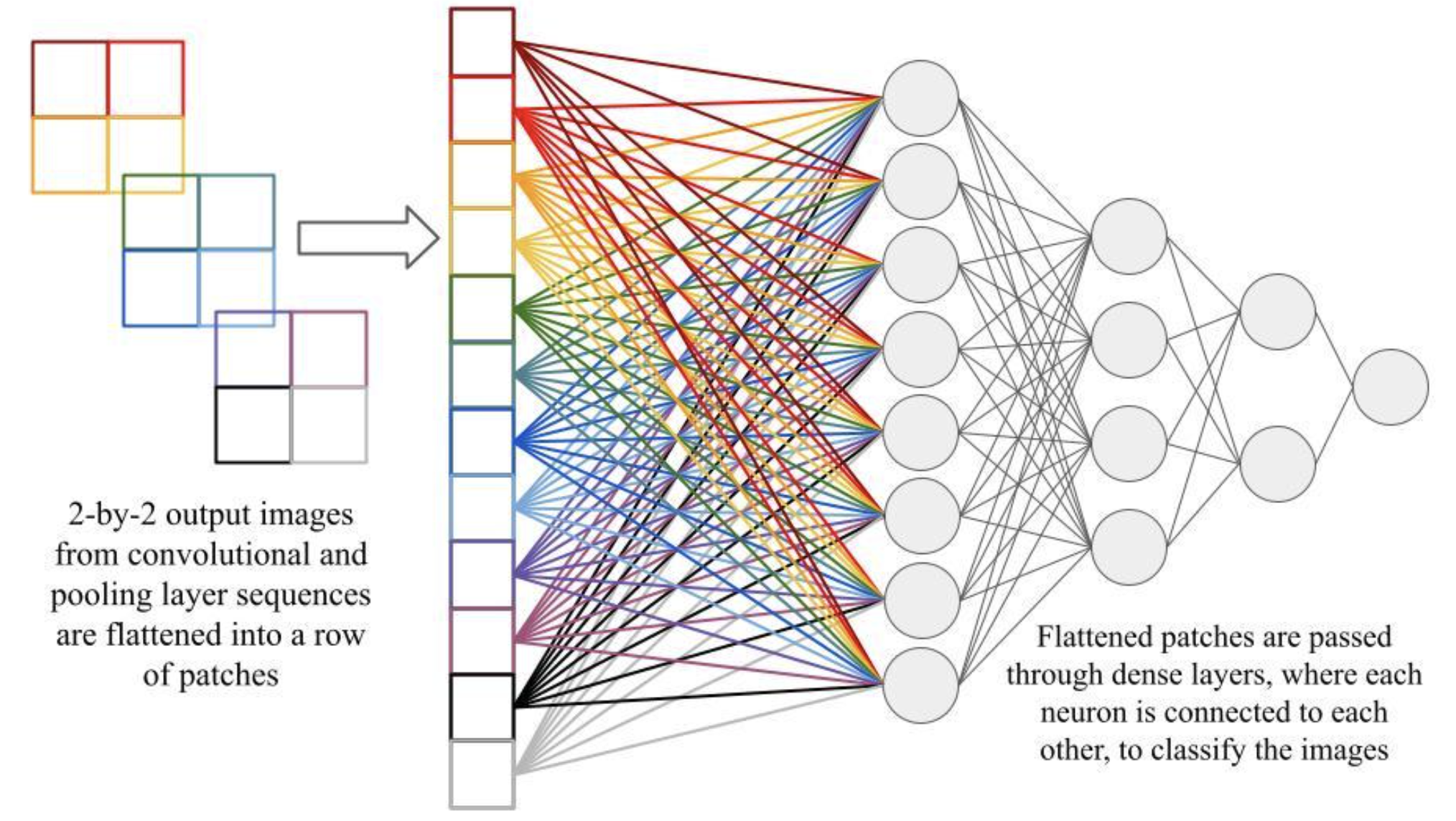}
         \caption{Dense Layers in CNN}
         \label{fig:2}
\end{figure}

Convolutional Neural Networks (CNNs) have gained prominence over the past decade due to their impressive modeling capabilities, making them optimal tools for image processing, computer vision and natural language processing\cite{li2021survey}. One of the key distinctions of a CNN is the presence of a convolutional layer that works by applying a convolutional filter to the given input, returning a filter map. It determines the output of neurons associated with input regions as a function of the weights and regions connected to input volumes. CNNs also include pooling layers, which reduce the dimensionality of the input by condensing multiple values in a feature map into one, through either an average or maximum function. These two layers help the model identify features of an image such as edges and textures. Then, through dense layers, which are linear combinations of multiple neurons, the CNNs learn the patterns and relationships to classify images. 
Figure \ref{fig:1} depicts an example of a convolutional layer followed by a pooling layer, starting with a 5-by-5 input image. As seen, the 3-by-3 filter iterates over the input image and performs the convolutional matrix dot product operation multiple times to return a 3-by-3 output. Then, maximum pooling is applied to the 3-by-3 image, where the maximum value of each subarray is identified. The final output is a 2-by-2 image. This process thus allows CNNs to transform input data to class specific scores which can then be used for classification applications. Figure \ref{fig:2} shows an example of dense layers in a CNN. The 2-by 2-feature maps are flattened into a singular row of neurons. Then, through a series of linear combinations, the neurons are combined together to identify the different patterns in the data and learn how to classify the images.

% \begin{figure}[ht]
%   \centering
%   \includegraphics[width=0.5\textwidth]{your-image-file.jpg} % Replace with your image file
%   \caption{Sample figure caption.}
%   \label{fig:sample1} 
% \end{figure}

In both convolutional and dense layers, an activation function is used at the end of the layer to introduce non linearity. The rectified linear unit (ReLu) is a commonly used activation function for the hidden layers, as shown in Eq. \ref{eq:eq1}, where $x$ is the output from the current layer and $R(x)$ is subsequently input to the next layer. The output layer also requires an activation function to generate the class probabilities for classification either through the sigmoid or softmax function. Binary class classification uses the sigmoid function $\sigma(x)$, where $x$ is simply the output of the network and $\sigma(x)$ is a value between 0 and 1 (Eq. \ref{eq:eq2}). If the value is above 0.50, the predicted class is the one associated with a 1, while a value below 0.50 indicates the predicted class is the one associated with a 0. Multi-class classification employs the softmax function  $S_i(x)$, where $x$ is the output vector of log probabilities and $S_i(x)$ represents the probability of the image belonging to class $x_i$ (Eq. \ref{eq:eq3}). 
\begin{equation}
    R(x) = max(0,x)
\label{eq:eq1}
\end{equation}
\begin{equation}
    \sigma(x) = \frac{1}{1+e^{-x}}
    \label{eq:eq2}
\end{equation}
\begin{equation}
    S_i(x) = \frac{e^{x_i}}{\sum_{j=1}^{K}e^{x_j}}
    \label{eq:eq3}
\end{equation}
\section{Methodology}
In this section, we detail the underlying CNN architecture that is used in combination with the proposed training methodology inspired from curriculum learning. 

\subsection{Iterative Misclassification Error Training (IMET)}

In many current medical datasets, the amount of data present is extremely limited- rare diseases especially have extremely few data points, which poses a challenge to neural network training. Through regular neural network training, extreme overfitting can occur due to the majority of the data belonging to one class, causing the model to solely learn the patterns present within the dominant class. As such, the overall performance of the neural network is significantly lowered, as it is unable to identify the rare class consistently. However, through efficient and planned strategies, these potential limitations of neural networks in medical imaging classification can be resolved. This challenge is addressed in this research through the proposed Iterative Misclassification Error Training (IMET) technique, which can significantly improve the training of a CNN.

The first potential solution to the problem of imbalanced data is equal class sampling. In this technique, the dataset is divided up into its classes and an equal number of data points are taken from each class as shown in the pseudo code in Algorithm 1. This method helps address problems with a very small minority class because it ensures that the model trains on an equal number of samples from each class,  preventing overfitting and alleviating imbalances in accuracies across classes. However, this technique may not be feasible in all cases, since, if there are very few samples of a minority class, then the amount of total training data would be extremely small and potentially insufficient to properly train the neural network. Additionally, the model may require more data for certain classes depending on image complexity.

Another approach could be a weighted sampling method \cite{provost2000machine}, which iteratively trains and updates the training data depending on the misclassification rate of the previous train. This technique helps the model better learn classes with lower accuracies by feeding it more examples from those classes, allowing the model to essentially update itself. The process starts with a random sample of images that will be used to initially train the model. Then, the model is evaluated and the misclassification rate for each class is found based on the number of incorrectly identified classes out of the whole train dataset. A new training dataset is subsequently created, where percent of images from each class is dependent on the class’s misclassification rate. The model is then re-trained using this new dataset, and the process of training through the misclassification error rate is repeated as necessary as explained in Algorithm 2. While this method seems promising, it can lead to the model overcorrecting due to a high misclassification rate for specific classes and contributing to the new training data of almost entirely those classes, resulting in overfitting and poorer accuracies on the other classes. As such, in the next round, the other classes are overcompensated with data, the model overfits one again, and the oscillation between accuracies continues. 

The proposed IMET technique incorporates elements from both the prior mentioned equal class sampling and the weighted sampling techniques, as shown in Algorithm 3. It starts with a train using equal class sampling, followed by an evaluation of the model's performance on a subsection of the train dataset. The subsection of the train datasets contains a random collection of an equal number of samples per class, where this number equals the size of the smallest class. This provides an idea of the classes that the model is classifying at lower rates by calculating the misclassification rate for each individual class. For the next train, 50 percent of the training data is gathered through the equal class sampling technique, where each class has the same amount of samples. The other 50 percent is calculated through the misclassification error rate as was done in weighted sampling. In this way, classes that were misclassified at high rates get compensated with more data while also ensuring that each class still receives a sufficient amount of data. The model is retrained using this data and the process is repeated multiple times. IMET alleviates the potential problem of lack of training data due to an extremely small minority class that is prevalent in equal class sampling. Moreover, ensuring that 50 percent of the data for the next train is obtained by equal class sampling also prevents the over-correction to specific classes often associated with weighted sampling.

\begin{algorithm}
\caption{Algorithm 1: Equal Sampling}
\label{alg:lad_generator1}
% \SetKwProg{generate}{Function \emph{generate}}{}{end}
\textbf{Input}: Data $d$, Number of classes $n$, data in each class $d_1, d_2 \dots d_{n-1}, d_n$, sample size $m$ for each class, DNN model $M$, number of times to retrain the model $N_{rep}$\\
\textbf{Output}: Updated train data $d_{T}$
% \textbf{Initialization}: Convert the data $d$ into $d_1, d_2 \dots d_{n-1}, d_n$ where n represents the number of classes within the dataset. 
\begin{algorithmic}[1]
\State $d_{norm} = Normalize(d)$ \label{Step_1}
\State $d_{train},d_{test} = Split\ train\ test(d_{norm})$
\For{each $i<=N_{rep}$}
\For{each  class $k<=n$}
        \State Randomly sample $m$ observations from $d_k$ in class $k$ as $d^k$.
\EndFor
\State Update the training sample $d_T = \bigcup_{k=1}^n d^k$ \label{Step_6}
\State Train model and update M on $d_T$ 
% \State Repeat Step \ref{Step_1} - Step \ref{Step_6}
\EndFor
\end{algorithmic}
\end{algorithm}

\begin{algorithm}
\caption{Algorithm 2: Weighted Sampling}
\label{alg:lad_generator2}
% \SetKwProg{generate}{Function \emph{generate}}{}{end}
\textbf{Input}: Data $d$, Number of classes $n$, data in each class $d_1, d_2 \dots d_{n-1}, d_n$, sample size $m$ for each class, DNN model $M$, number of times to retrain the model $N_{rep}$, inital sample size $k$\\
\textbf{Output}: Updated train data $d_{T}$
% \textbf{Initialization}: Convert the data $d$ into $d_1, d_2 \dots d_{n-1}, d_n$ where n represents the number of classes within the dataset. 
\begin{algorithmic}[1]
\State $d_{norm} = Normalize(d)$ 
\State $d_{train},d_{test} = Split\ train\ test(d_{norm})$
\State Sample $k$ data points randomly from each class as $d_T$
\State Train model M on $d_T$ 
\For{each $i<=N_{rep}$}
\State Compute the model M predictions on $d_{train}$ and the missclassified samples in each class given by $m_1,\dots,m_n$ respectively \label{alg2_start2}
\State Compute the counts of misclassified sampled from each class, $w_i = |m_i|, \forall i \in 1,\dots,n$
\State Compute percentage weights $p_k = \frac{w_k}{\Sigma_{k=1}^n w_k}*100,\  \forall k\in 1,\dots,n$
\For{each  class $k<=n$}
        \State Randomly sample $p_k$\% observations from $m_k$ in class $k$ as $d^k$.
\EndFor
\State Update the training sample $d_T = \bigcup_{k=1}^n d^k$ \label{alg2_end}
\State Train model and update M on $d_T$ 
% \State Repeat Step \ref{alg2_start} - Step \ref{alg2_end}
\EndFor
\end{algorithmic}
\end{algorithm}

\begin{algorithm}
\caption{Algorithm 3: Iterative based Misclassification Error Training (IMET)}
\label{alg:lad_generator3}
% \SetKwProg{generate}{Function \emph{generate}}{}{end}
\textbf{Input}: Data $d$, Number of classes $n$, data in each class $d_1, d_2 \dots d_{n-1}, d_n$, sample size $m$ for each class, DNN model $M$, number of times to retrain the model $N_{rep}$ , inital sample size $k$\\
\textbf{Output}: Updated train data $d_{T}$\\
\textbf{Initialization}: Using the smallest class size $m=min(|d_1|,\dots, |d_n|)$, compute a sub-sample of train data $d_{sub}$ which is a collection of $m$ random samples from each class
\begin{algorithmic}[1]
\State $d_{norm} = Normalize(d)$ 
\State $d_{train},d_{test} = Split\ train\ test(d_{norm})$
\State Sample $k$ data points randomly from each class as $d_T$
\State Train model M on $d_T$ 
\For{each $i<=N_{rep}$}
\State Compute the model M predictions on $d_{sub}$ and the missclassified samples in each class given by $m_1,\dots,m_n$ respectively \label{alg2_start3}
\State Compute the counts of misclassified sampled from each class, $w_i = |m_i|, \forall i \in 1,\dots,n$
\State Compute percentage weights $p_k = \frac{w_k}{\Sigma_{k=1}^n w_k}*100,\  \forall k\in 1,\dots,n$
\For{each  class $k<=n$}
        \State Randomly sample $p_k$\% observations from $m_k$ in class $k$ as $d^k$.
        \State Randomly sample m observations from $d_k$ in class k and add to $d^k$.
\EndFor
\State Update the training sample $d_T = \bigcup_{k=1}^n d^k$ \label{alg2_end}
\State Train model and update M on $d_T$ 
% \State Repeat Step \ref{alg2_start} - Step \ref{alg2_end}
\EndFor
\end{algorithmic}
\end{algorithm}

\subsection{Proposed CNN Architecture}
\begin{figure}
    \centering
    \includegraphics[width=1\linewidth]{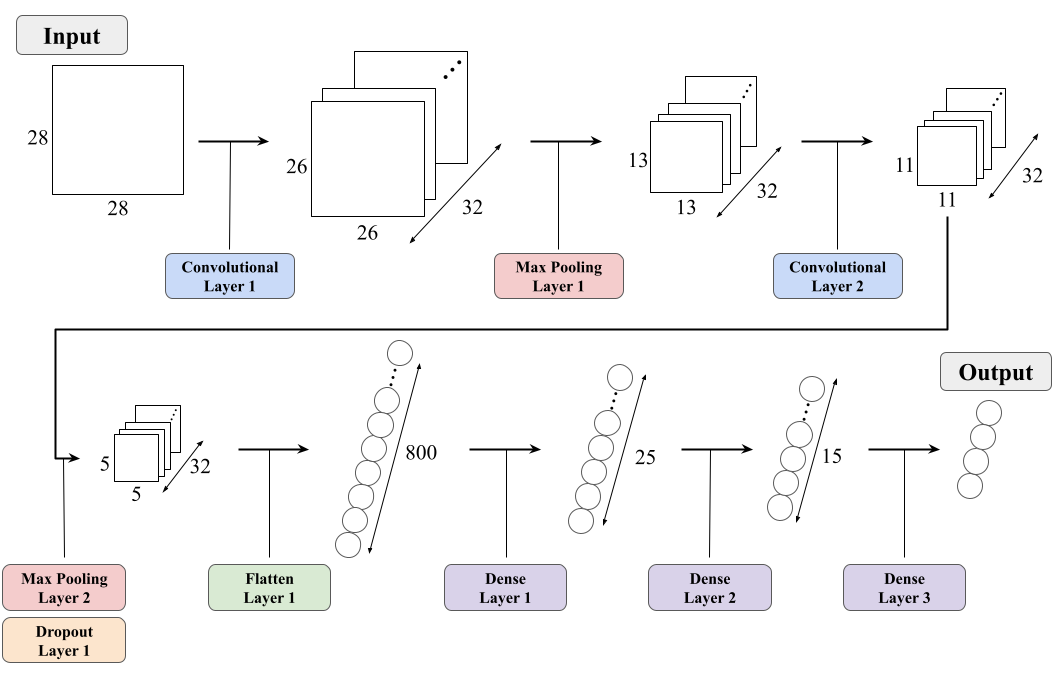}
    \caption{CNN model architecture, highlighting the output after each layer.}
    \label{fig:fig5}
\end{figure}

Figure \ref{fig:fig5} depicts the CNN architecture used in this research. It consists of 9 layers, starting with a grayscale 28-by-28 input image and ending with an output of the predicted class for the input image. The model was built using Tensorflow using the Google Colab Python interface. The initial layer of the architecture is a 2D-convolutional layer that transforms the original 28- by-28 pixel image into 32 26-by-26 filtermaps. After the convolution layer, max pooling was used to shrink the dimensionality of the feature maps to 13-by-13. Furthermore, another round of the 2D-convolutional filter followed by max pooling was applied to further reduce the dimensions to 5-by 5-by-32. Next, in order to reduce the chances of potential overfitting during training, a dropout layer that freezes half the neurons was implemented. Then, the filtermaps were flattened into a single row of 800 neurons, after which multiple dense layers were interconnected to capture patterns of different classes. In the final layer, the softmax function was used to generate the probability distributions of the multi-class OCTMNIST images, while the sigmoid function was used to generate the probability distributions of the binary class PneumoniaMNIST images. The ReLU activation function was used throughout each convolutional and dense layer. Additionally, each technique trained with 100 epochs and used the adaptive moment estimator (ADAM) with an initial learning rate of 0.001 and weight decay as the optimizer to be consistent with the MedMNIST models. 

\section{Results}
This section presents the outcomes of the experiments conducted to evaluate the IMET training approach. We detail the datasets used as benchmarks, performance metrics, and comparisons with baseline models to demonstrate the contributions of our IMET training. The results from the IMET technique were evaluated using the Receiving Operator Curve (ROC)-Area Under the Curve (AUC), Confusion Matrices and comparison with the benchmark MedMNIST models. 

\subsection{Data Sources}
This research used two datasets from the MedMNIST collection. MedMNIST v2 \cite{yang2023medmnist} is a publicly available collection of standardized medical datasets that are commonly used for biomedical image analysis. The two MedMNIST datasets specifically used in this research and their associated characteristics are shown in Table \ref{tab:tab1}. The datasets are all pre-processed into train, test, and validation sets which include 28-by-28 images, and can be used in either grayscale or RGB form. 

\begin{table}[]
    \centering
    \begin{tabular}{|c|c|c|c|c|}
        \hline\\
       \textbf{ Dataset} & \textbf{Modality} & \textbf{Size} & \textbf{Classes} & \textbf{Color}\\
        \hline \\
        OCTMNIST & Retinal OCT & 109,309 & 4 & greyscale \\
        \hline \\
         PneumoniaMNIST & Chest X-Ray & 5,856 & 2 & greyscale \\
         \hline
    \end{tabular}
    \caption{Data description}
    \label{tab:tab1}
\end{table}

\begin{figure}
    \centering
    \includegraphics[width=0.95\linewidth]{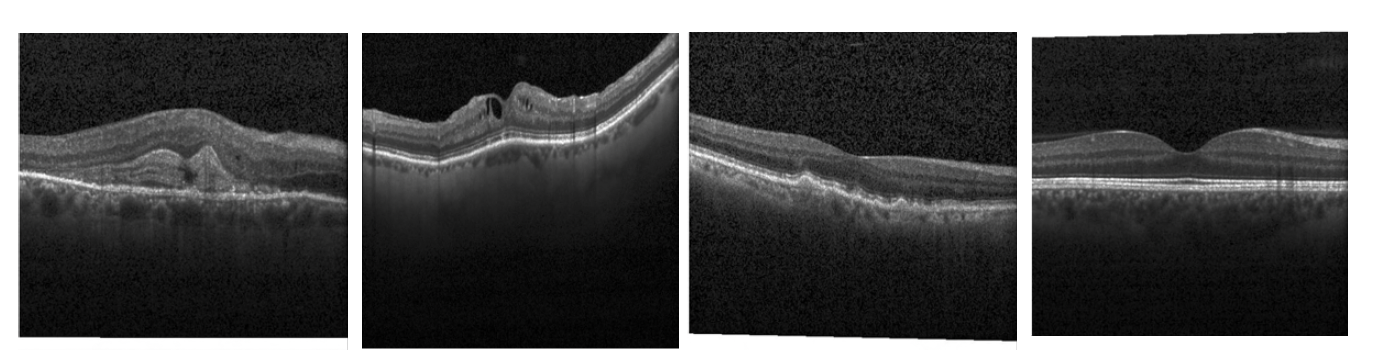}
    \caption{Sample images of the categories from OCTMNIST dataset. From right to left: CNV, DME, Drusen, and normal retina. }
    \label{fig:fig3}
\end{figure}

OCTMNIST has been compiled from a prior dataset of 109,309 optical coherence tomography (OCT) images from 4,686 patients for retinal diseases\cite{kermany2018identifying}. The dataset consists of 4 categories: 37,455 images of  Choroidal Neovascularization (CNV) identified by label 0, 11598 images of Diabetic Macular Edema identified by label 1, 8866 images of Drusen identified by label 2, and 51390 images of normal retina identified by label 3 (Figure \ref{fig:fig3}). Choroidal Neovascularization (CNV) is characterized by the growth of new blood vessels into the sub-retinal pigment epithelium (sub-RPE) or subretinal space leading to vision loss. Diabetic Macular Edema (DME) results from fluid accumulation in the central part of the retina or macula that affects the fovea where vision is the sharpest. DME is a complication from diabetes resulting in vision loss. Drusen are yellow deposits made up of lipids and proteins and are present under the retina. Drusen, when present in large numbers, is an early indicator of age-related macular degeneration. The OCTMNIST data is split into 97,477 training, 10,832 validation and 1,000 test images.

\begin{figure}
    \centering
    \includegraphics[width=0.95\linewidth]{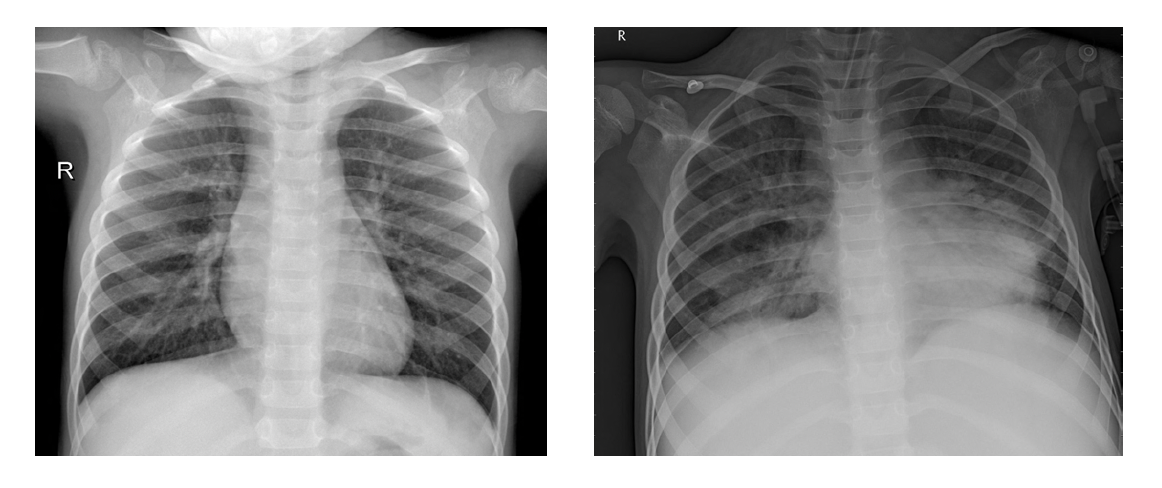}
    \caption{X-Ray images of a healthy lung (left) and a pneumonia-infected lung (right) from the PneumoniaMNIST dataset.}
    \label{fig:fig4}
\end{figure}

The PneumoniaMNIST dataset has also been compiled from a prior dataset of 5,856 pediatric chest X-Ray images \cite{kermany2018identifying}. Of these images, 1,349 depict healthy lungs, while 3,883 display lungs affected by pneumonia (Figure \ref{fig:fig4}). The dataset was further split into 4,708 training, 524 validation and  624 test images.

\subsection{Model Evaluation}

In this research accuracy, precision, recall, and the F1-score were used as the primary measures of model performance. Accuracy is a measure of the number of samples that were correctly classified by the model out of all data points (Eq. \ref{eq:eq4}). Precision measures the number of examples which the model correctly identifies as the positive class out of all the examples which the model determined to be the positive class (Eq. \ref{eq:eq5}) . Recall measures how often the model predicted the positive class out of all the examples that were actually of the positive class (Eq. \ref{eq:eq6}). The F1-score. is the harmonic mean of precision and recall (Eq \ref{eq:eq7}).

\begin{equation}
    accuracy = \frac{TP + TN}{TP+TN+FP+FN}
    \label{eq:eq4}
\end{equation}
\begin{equation}
    precision = \frac{TP}{TP+FP}
    \label{eq:eq5}
\end{equation}
\begin{equation}
    recall = \frac{TP}{TP+FN}
    \label{eq:eq6}
\end{equation}
\begin{equation}
    F1 = 2*\frac{precision*recall}{precision+recall}
    \label{eq:eq7}
\end{equation}

where, TP refers to true positives, FP refers to false positives, TN refers to true negatives, and FN refers to false negatives. True positive is defined as the number of examples of a class that were correctly identified to the positive class while true negative is the number of examples correctly classified to the negative class. False negative is the number of examples where the model incorrectly predicts the positive class and the false negative is the number of examples incorrectly predicted as the negative class.

\subsection{ROC-AUC Curves}

\begin{figure*}
    \centering
    \includegraphics[width=\linewidth]{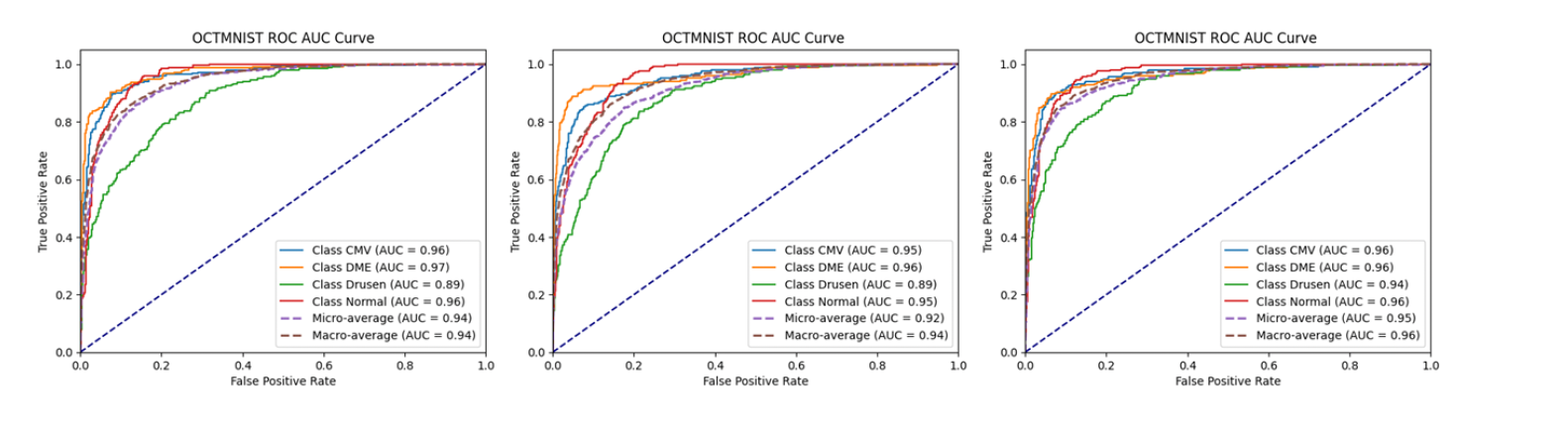}
    \caption{OCTMNIST ROC curves. Left to right: Equal Class Sampling, Weighted Sampling, IMET}
    \label{fig:fig6}
\end{figure*}

The ROC is the plot of the false positive rate on the x axis and the true positive rate on the y axis over different threshold values. For multi-classification, an ROC curve can be generated for each class in comparison to all the other classes. Additionally, a micro average ROC can be calculated where each data point has equal weight or a macro average where each class has the same weightage in calculating the ROC. AUC is simply the Area Under the Curve of a ROC plot and ranges between 0 and 1, where a completely random classifier would have an AUC of 0.5 AUC is an important metric as it is calculated across multiple thresholds and is generally not as affected by class imbalance as accuracy is. 
The ROC and AUC for the OCTMNIST dataset for the different training techniques are shown in Figure \ref{fig:fig6}. The blue line is calculated by taking CNV as one class and meshing the remaining classes together into another class. Likewise, the orange line represents the same for DME, the green line for Drusen and the red line for the normal retina images. The dotted purple line is the micro average which calculates the AUC by adding the predictions for all the classes at once. This means each datapoint has equal weightage when determining the AUC. Whereas, the macro average denoted by the dotted brown line calculated the metric for each class and then averaged them so that each class had an equal weightage in determining the AUC. It can be observed that the AUC for both the macro and micro averaged curves was the highest for the IMET technique.

\begin{figure*}
    \centering
    \includegraphics[width=0.95\linewidth]{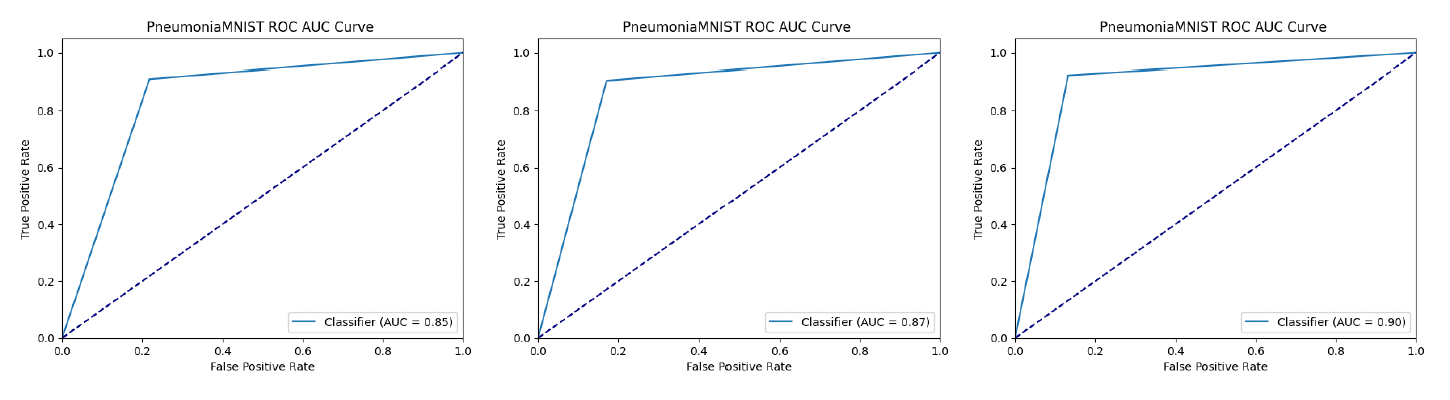}
    \caption{PneumoniaMNIST ROC curves. Left to right: Equal Class Sampling, Weighted Sampling, IMET}
    \label{fig:fig7}
\end{figure*}

Similarly, the ROC-AUC curve was calculated for the PneumoniaMNIST dataset. However, since the dataset has only two classes, only one curve was calculated as shown in Figure \ref{fig:fig7}. The highest AUC score was obtained by IMET with a value of 0.90.

\subsection{Confusion Matrices}

\begin{figure*}
    \centering
    \includegraphics[width=0.95\linewidth]{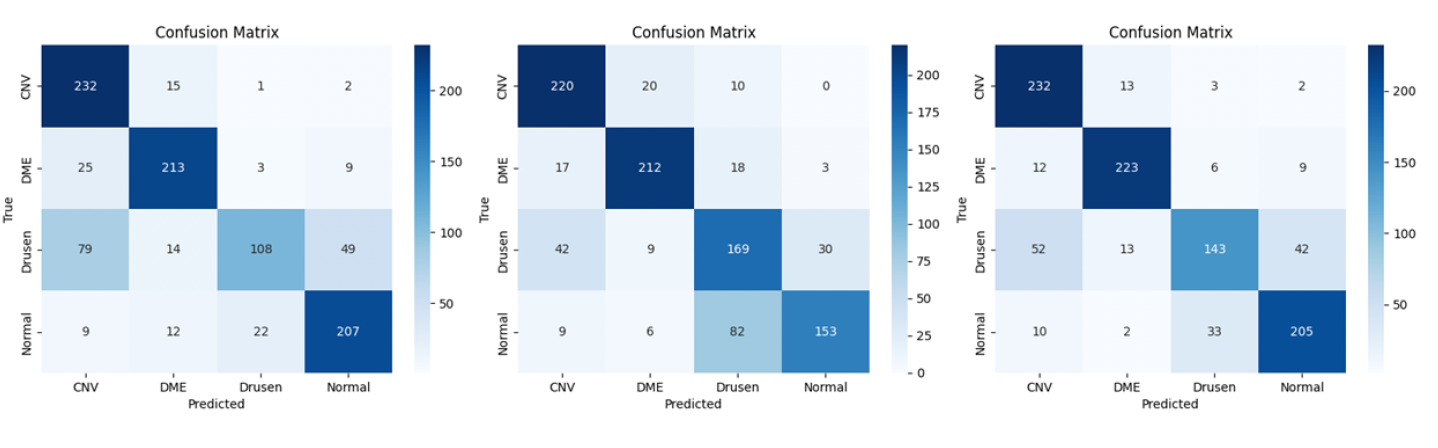}
    \caption{OCTMNIST Confusion Matrix. Left to right: Equal Class Sampling, Weighted Sampling, IMET}
    \label{fig:fig8}
\end{figure*}

A confusion matrix is a N x N matrix that evaluates the performance of a classification model, where N is the total number of target classes. It provides the number of true positives, true negatives, false positives, and false negatives in a classification model, thereby identifying mis-classifications and helping with improving the predictive accuracy. Overall, it provides an overview of the model's effectiveness and the kind of errors it may likely make.
The OCTMNIST dataset was trained using 3 techniques (equal sampling, weighted sampling, and IMET), and each trained model was then tested on a test dataset containing 1000 images. The confusion matrices generated from the testing are shown in Figure \ref{fig:fig8}. The x-axis of the confusion matrices displays the class that was predicted by the trained model while the y-axis displays the actual class. The diagonal of the confusion matrix presents the number of all testing images that were correctly classified, which was 760, 754, and 803 for equal class, weighted, and IMET techniques, respectively. Additionally, 18 images for the CNV class, 37 images of DME, 142 images of Drusen and 43 images of a normal retina were incorrectly classified using the equal class technique. Similarly, 30 images for the CNV class, 38 images of DME, 81 images of drusen and 97 images of a normal retina were incorrectly classified using the weighted sampling technique. Finally, 18 images for the CNV class, 27 images of DME, 107 images of drusen and 45 images of a normal retina were incorrectly classified using the IMET technique. Other accuracy metrics (precision, accuracy, recall and F1 score) that were calculated from the confusion matrices for the three techniques for the OCTMNIST dataset are presented in Table \ref{tab:tab2}.

\begin{figure*}
    \centering
    \includegraphics[width=0.95\linewidth]{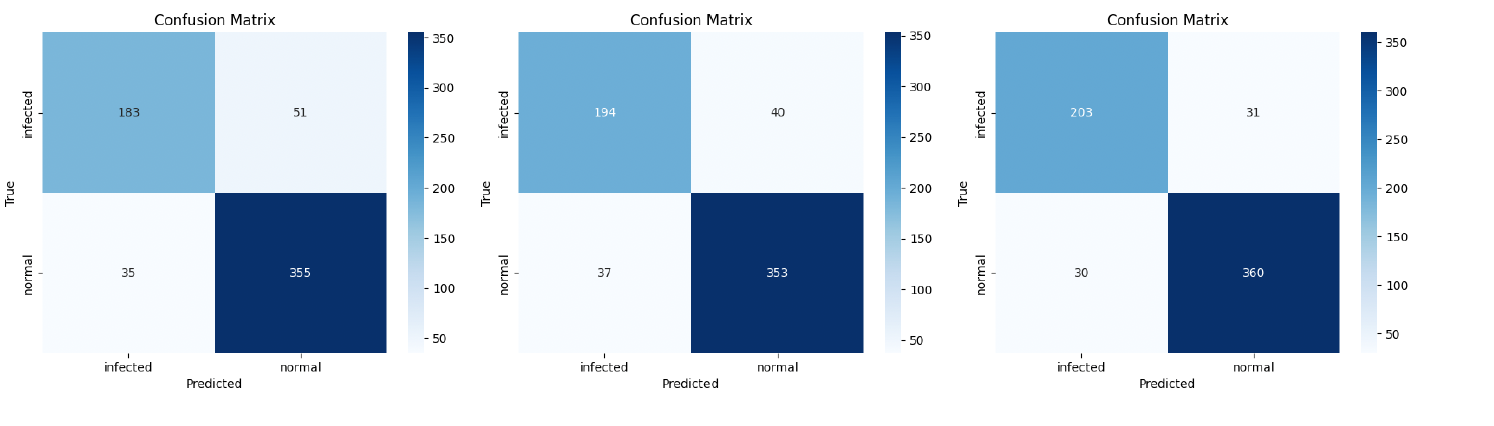}
    \caption{PneumoniaMNIST Confusion Matrix. Left to right: Equal Class Sampling, Weighted Sampling, IMET}
    \label{fig:fig9}
\end{figure*}

In a similar fashion, confusion matrices were generated for the testing dataset for PneumoniaMNIST dataset, which contained 624 images of either infected or normal lung x-rays (Figure \ref{fig:fig9}). The diagonal of the confusion matrix presents the number of all testing examples that were correctly classified, which were 538, 547, and 563 for equal class, weighted, and IMET techniques, respectively. Additionally, 51 images for the infected classes and 35 images of a normal chest x-ray were incorrectly classified for the equal class technique. Likewise, 40 images for the infected classes and 37 images of a normal chest x-ray were incorrectly classified for the weighted sampling technique. Finally, 31 images for the infected classes and 30 images of a normal chest x-ray were incorrectly classified for the IMET technique. Other accuracy metrics (precision, accuracy, recall and F1 score) that were calculated from the confusion matrices for the three techniques for the PneumoniaMNIST dataset are presented in Table \ref{tab:tab3}.

\subsection{Comparisons with Benchmark Models}

The MedMNIST collection consists of benchmark results from various residual networks (ResNet), auto-sklearn, and Google's auto-ml. However for this research, only the results from the ResNet models were considered as the other techniques are composed of vastly different architectures and not directly comparable. The results from the IMET technique along with the results from equal class sampling and weighted sampling were compared to the results from the OCTMNIST ResNet benchmark models. The performance of the different techniques for various metrics on the OCTMNIST dataset are shown in Table \ref{tab:tab2}. As seen from the table, IMET displayed the highest accuracy of 80.6\%, followed by weighted sampling (79.4\%) even with a significantly lower parameter count (around 366 times lower) and lower number of training samples (87,000 in comparison to the 97,000 used by the ResNets). AUC scores were comparable amongst the ResNet models and IMET. IMET also provided the highest precision, recall, and F1 scores of 0.804, 0.806, and 0.802 amongst the techniques used in this research (precision, recall and F1 measure were not reported for benchmark models).

\begin{table*}
    \caption{OCTMNIST Results}
    \label{tab:tab2}
    \begin{tabular}{|c|c|c|c|c|c|c|c|}
        \hline
        Model & Parameters & Samples & Accuracy & AUC & Precision & Recall & F1 \\
        \hline
        ResNet-18 (28) & 11,169,732 & 97,477 & 0.743 & 0.943 & N/A & N/A & N/A \\
        ResNet-18 (224) & 11,169,732 & 97,477 & 0.763 & 0.958 & N/A & N/A & N/A \\
        ResNet-50 (28) & 23,507,396 & 97,477 & 0.762 & 0.952 & N/A & N/A & N/A \\
        ResNet-50 (224) & 23,507,396 & 97,477 & 0.776 & 0.958 & N/A & N/A & N/A \\
        Baseline CNN & 30,047 & 97,477 & 0.693 & 0.951 & 0.787 & 0.693 & 0.651 \\
        Equal Class Sampling & 30,047 & 31,016 & 0.76 & 0.944 & 0.773 & 0.76 & 0.747 \\
        Weighted Sampling & 30,047 & 63,000 & 0.754 & 0.935 & 0.798 & 0.754 & 0.753 \\
        IMET & 30,047 & 87,000 & 0.803 & 0.96 & 0.804 & 0.803 & 0.8 \\
        \hline
    \end{tabular}
\end{table*}

\begin{table*}[!htbp]
    \caption{PneumoniaMNIST Results}
    \label{tab:tab3}
    \begin{tabular}{|c|c|c|c|c|c|c|c|}
    \hline
    Model & Parameters & Samples & Accuracy & AUC & Precision & Recall & F1 \\
    \hline
ResNet-18 (28)	&	11,168,706	&	4,708	&	0.854	&	0.944	&	N/A	&	N/A	&	N/A	\\
ResNet-18 (224)	&	11,168,706	&	4,708	&	0.864	&	0.956	&	N/A	&	N/A	&	N/A	\\
ResNet-50 (28)	&	23,503,298	&	4,708	&	0.854	&	0.948	&	N/A	&	N/A	&	N/A	\\
ResNet-50 (224)	&	23,503,298	&	4,708	&	0.884	&	0.962	&	N/A	&	N/A	&	N/A	\\
Baseline CNN	&	30,047	&	4708	&	0.843	&	0.818	&	0.825	&	0.872	&	0.848	\\
Equal Class Sampling	&	30,047	&	2428	&	0.862	&	0.846	&	0.874	&	0.901	&	0.892	\\
Weighted Sampling	&	30,047	&	4700	&	0.877	&	0.867	&	0.884	&	0.877	&	0.876	\\
IMET	&	30,047	&	2800	&	0.902	&	0.895	&	0.901	&	0.902	&	0.902	\\
\hline
    \end{tabular}
\end{table*}

Similarly, Table \ref{tab:tab3} shows the performance of the different techniques on the PneumoniaMNIST dataset for various metrics in comparison to the ResNet benchmark models. As seen from the table, IMET has an accuracy of 90.2\%, weighted sampling an accuracy of 87.7\% and equal class sampling an accuracy of 86.2\%. Moreover, IMET had the highest accuracy, precision, and F1 scores amongst the techniques used in this research and required only 2,800 training samples, compared to 4,700 training samples by the ResNet models.

\section{Conclusion and Future Work}
Extensive training on medical datasets enables the model to recognize different image classes based on potential diseases and ailments, resulting in the model being able to accurately predict diagnoses solely based on images. However, the lack of high-quality data in medical sciences is challenging due to privacy concerns, data sharing restrictions, and the labor-intensive process of manual annotation. This challenge is exacerbated by the fact that many healthcare organizations lack the necessary datasets, and certain diseases exhibit imbalanced data. Additionally, the computational demands of training complex deep learning models pose a barrier to model adaption, as it requires substantial computing power and expertise. 
This research demonstrated that, through the use of novel training techniques based on misclassified observations, improved performance on classifying biomedical images can be achieved. The proposed Iterative Misclassification Error Training (IMET) technique used a smaller CNN with only 30,047 parameters (roughly 366 times smaller than the ResNet-18 and 765 times smaller than ResNet-50) to iteratively update the batch of training data based on the misclassification error from each class, successfully outperforming the OCTMNIST and PneumoniaMNIST benchmark models. It used only 87,000 samples of training data and achieved an accuracy of 80.3\% for the OCTMNIST data compared to the 77.6\% accuracy of the ResNet-50 (224) that used 97,477 training samples. For PneumoniaMNIST, IMET obtained an accuracy of 90.2\% with only 2800 training samples as opposed to an accuracy of 88.6\% for the ResNet-50 (224) that used 4708 training samples. This novel process of feeding data through the IMET technique during a neural network training can boost performance of a model and achieve better accuracy with considerably fewer training data or class imbalance. However, this technique has only been tested on medical images, so its performance on other datasets is unknown and more extensive testing will be required to determine its efficacy on other data formats and domains.

The IMET technique developed in this research is not domain specific and can be used in other applications. Testing this technique on different architectures such as transformers and large language models is another potential step to see if this technique can help alleviate the requirement of large amounts of data in training. Additionally, implementing this technique using multi-modal datasets (data that combines text with images), which are common in the medical field, can be another future opportunity.

\bibliographystyle{acm}
\bibliography{main}

\begin{thebibliography}{10}

\bibitem{abramoff2016improved}
{\sc Abr{\`a}moff, M.~D., Lou, Y., Erginay, A., Clarida, W., Amelon, R., Folk, J.~C., and Niemeijer, M.}
\newblock Improved automated detection of diabetic retinopathy on a publicly available dataset through integration of deep learning.
\newblock {\em Investigative ophthalmology \& visual science 57}, 13 (2016), 5200--5206.

\bibitem{alazab2020covid}
{\sc Alazab, M., Awajan, A., Mesleh, A., and Alhyari, S.}
\newblock Covid-19 prediction and detection using deep learning.
\newblock {\em International Journal of Computer Information Systems and Industrial Management Applications 12\/} (2020), 14--14.

\bibitem{bengio2009curriculum}
{\sc Bengio, Y., Louradour, J., Collobert, R., and Weston, J.}
\newblock Curriculum learning.
\newblock In {\em Proceedings of the 26th annual international conference on machine learning\/} (2009), pp.~41--48.

\bibitem{bi2019effective}
{\sc Bi, X.-a., Cai, R., Wang, Y., and Liu, Y.}
\newblock Effective diagnosis of alzheimer’s disease via multimodal fusion analysis framework.
\newblock {\em Frontiers in genetics 10\/} (2019), 976.

\bibitem{chai2023efficient}
{\sc Chai, C., Wang, J., Tang, N., Yuan, Y., Liu, J., Deng, Y., and Wang, G.}
\newblock Efficient coreset selection with cluster-based methods.
\newblock In {\em Proceedings of the 29th ACM SIGKDD Conference on Knowledge Discovery and Data Mining\/} (2023), pp.~167--178.

\bibitem{cirecsan2013mitosis}
{\sc Cire{\c{s}}an, D.~C., Giusti, A., Gambardella, L.~M., and Schmidhuber, J.}
\newblock Mitosis detection in breast cancer histology images with deep neural networks.
\newblock In {\em Medical Image Computing and Computer-Assisted Intervention--MICCAI 2013: 16th International Conference, Nagoya, Japan, September 22-26, 2013, Proceedings, Part II 16\/} (2013), Springer, pp.~411--418.

\bibitem{cruz2014automatic}
{\sc Cruz-Roa, A., Basavanhally, A., Gonz{\'a}lez, F., Gilmore, H., Feldman, M., Ganesan, S., Shih, N., Tomaszewski, J., and Madabhushi, A.}
\newblock Automatic detection of invasive ductal carcinoma in whole slide images with convolutional neural networks.
\newblock In {\em Medical imaging 2014: Digital pathology\/} (2014), vol.~9041, SPIE, p.~904103.

\bibitem{feldman2020core}
{\sc Feldman, D.}
\newblock Core-sets: Updated survey.
\newblock {\em Sampling techniques for supervised or unsupervised tasks\/} (2020), 23--44.

\bibitem{guo2018curriculumnet}
{\sc Guo, S., Huang, W., Zhang, H., Zhuang, C., Dong, D., Scott, M.~R., and Huang, D.}
\newblock Curriculumnet: Weakly supervised learning from large-scale web images.
\newblock In {\em Proceedings of the European conference on computer vision (ECCV)\/} (2018), pp.~135--150.

\bibitem{hacohen2019power}
{\sc Hacohen, G., and Weinshall, D.}
\newblock On the power of curriculum learning in training deep networks.
\newblock In {\em International conference on machine learning\/} (2019), PMLR, pp.~2535--2544.

\bibitem{han2005borderline}
{\sc Han, H., Wang, W.-Y., and Mao, B.-H.}
\newblock Borderline-smote: a new over-sampling method in imbalanced data sets learning.
\newblock In {\em International conference on intelligent computing\/} (2005), Springer, pp.~878--887.

\bibitem{hu2023dual}
{\sc Hu, J., Shen, A., Qiao, X., Zhou, Z., Qian, X., Zheng, Y., Bao, J., Wang, X., and Dai, Y.}
\newblock Dual attention guided multiscale neural network trained with curriculum learning for noninvasive prediction of gleason grade group from mri.
\newblock {\em Medical Physics 50}, 4 (2023), 2279--2289.

\bibitem{jiang2014easy}
{\sc Jiang, L., Meng, D., Mitamura, T., and Hauptmann, A.~G.}
\newblock Easy samples first: Self-paced reranking for zero-example multimedia search.
\newblock In {\em Proceedings of the 22nd ACM international conference on Multimedia\/} (2014), pp.~547--556.

\bibitem{katharopoulos2018not}
{\sc Katharopoulos, A., and Fleuret, F.}
\newblock Not all samples are created equal: Deep learning with importance sampling.
\newblock In {\em International conference on machine learning\/} (2018), PMLR, pp.~2525--2534.

\bibitem{kermany2018identifying}
{\sc Kermany, D.~S., Goldbaum, M., Cai, W., Valentim, C.~C., Liang, H., Baxter, S.~L., McKeown, A., Yang, G., Wu, X., Yan, F., et~al.}
\newblock Identifying medical diagnoses and treatable diseases by image-based deep learning.
\newblock {\em cell 172}, 5 (2018), 1122--1131.

\bibitem{killamsetty2021retrieve}
{\sc Killamsetty, K., Zhao, X., Chen, F., and Iyer, R.}
\newblock Retrieve: Coreset selection for efficient and robust semi-supervised learning.
\newblock {\em Advances in neural information processing systems 34\/} (2021), 14488--14501.

\bibitem{kropp2023diabetic}
{\sc Kropp, M., Golubnitschaja, O., Mazurakova, A., Koklesova, L., Sargheini, N., Vo, T.-T. K.~S., de~Clerck, E., Polivka~Jr, J., Potuznik, P., Polivka, J., et~al.}
\newblock Diabetic retinopathy as the leading cause of blindness and early predictor of cascading complications—risks and mitigation.
\newblock {\em Epma Journal 14}, 1 (2023), 21--42.

\bibitem{lapedriza2013all}
{\sc Lapedriza, A., Pirsiavash, H., Bylinskii, Z., and Torralba, A.}
\newblock Are all training examples equally valuable?
\newblock {\em arXiv preprint arXiv:1311.6510\/} (2013).

\bibitem{li2021survey}
{\sc Li, Z., Liu, F., Yang, W., Peng, S., and Zhou, J.}
\newblock A survey of convolutional neural networks: analysis, applications, and prospects.
\newblock {\em IEEE transactions on neural networks and learning systems 33}, 12 (2021), 6999--7019.

\bibitem{liu2014early}
{\sc Liu, S., Liu, S., Cai, W., Pujol, S., Kikinis, R., and Feng, D.}
\newblock Early diagnosis of alzheimer's disease with deep learning.
\newblock In {\em 2014 IEEE 11th international symposium on biomedical imaging (ISBI)\/} (2014), IEEE, pp.~1015--1018.

\bibitem{moser2025coreset}
{\sc Moser, B.~B., Shanbhag, A.~S., Frolov, S., Raue, F., Folz, J., and Dengel, A.}
\newblock A coreset selection of coreset selection literature: Introduction and recent advances.
\newblock {\em arXiv preprint arXiv:2505.17799\/} (2025).

\bibitem{padhy2019artificial}
{\sc Padhy, S.~K., Takkar, B., Chawla, R., and Kumar, A.}
\newblock Artificial intelligence in diabetic retinopathy: A natural step to the future.
\newblock {\em Indian journal of ophthalmology 67}, 7 (2019), 1004--1009.

\bibitem{pratt2016convolutional}
{\sc Pratt, H., Coenen, F., Broadbent, D.~M., Harding, S.~P., and Zheng, Y.}
\newblock Convolutional neural networks for diabetic retinopathy.
\newblock {\em Procedia computer science 90\/} (2016), 200--205.

\bibitem{provost2000machine}
{\sc Provost, F.}
\newblock Machine learning from imbalanced data sets 101.
\newblock In {\em Proceedings of the AAAI’2000 workshop on imbalanced data sets\/} (2000), vol.~68, AAAI Press, pp.~1--3.

\bibitem{russakovsky2015imagenet}
{\sc Russakovsky, O., Deng, J., Su, H., Krause, J., Satheesh, S., Ma, S., Huang, Z., Karpathy, A., Khosla, A., Bernstein, M., et~al.}
\newblock Imagenet large scale visual recognition challenge.
\newblock {\em International journal of computer vision 115\/} (2015), 211--252.

\bibitem{santiago2021low}
{\sc Santiago, C., Barata, C., Sasdelli, M., Carneiro, G., and Nascimento, J.~C.}
\newblock Low: Training deep neural networks by learning optimal sample weights.
\newblock {\em Pattern recognition 110\/} (2021), 107585.

\bibitem{soviany2022curriculum}
{\sc Soviany, P., Ionescu, R.~T., Rota, P., and Sebe, N.}
\newblock Curriculum learning: A survey.
\newblock {\em International Journal of Computer Vision 130}, 6 (2022), 1526--1565.

\bibitem{sun2018integrating}
{\sc Sun, D., Li, A., Tang, B., and Wang, M.}
\newblock Integrating genomic data and pathological images to effectively predict breast cancer clinical outcome.
\newblock {\em Computer methods and programs in biomedicine 161\/} (2018), 45--53.

\bibitem{venugopalan2021multimodal}
{\sc Venugopalan, J., Tong, L., Hassanzadeh, H.~R., and Wang, M.~D.}
\newblock Multimodal deep learning models for early detection of alzheimer’s disease stage.
\newblock {\em Scientific reports 11}, 1 (2021), 3254.

\bibitem{wang2012machine}
{\sc Wang, S., and Summers, R.~M.}
\newblock Machine learning and radiology.
\newblock {\em Medical image analysis 16}, 5 (2012), 933--951.

\bibitem{wang2021survey}
{\sc Wang, X., Chen, Y., and Zhu, W.}
\newblock A survey on curriculum learning.
\newblock {\em IEEE transactions on pattern analysis and machine intelligence 44}, 9 (2021), 4555--4576.

\bibitem{wei2015submodularity}
{\sc Wei, K., Iyer, R., and Bilmes, J.}
\newblock Submodularity in data subset selection and active learning.
\newblock In {\em International conference on machine learning\/} (2015), PMLR, pp.~1954--1963.

\bibitem{wernick2010machine}
{\sc Wernick, M.~N., Yang, Y., Brankov, J.~G., Yourganov, G., and Strother, S.~C.}
\newblock Machine learning in medical imaging.
\newblock {\em IEEE signal processing magazine 27}, 4 (2010), 25--38.

\bibitem{xia2022moderate}
{\sc Xia, X., Liu, J., Yu, J., Shen, X., Han, B., and Liu, T.}
\newblock Moderate coreset: A universal method of data selection for real-world data-efficient deep learning.
\newblock In {\em The Eleventh International Conference on Learning Representations\/} (2022).

\bibitem{yang2023medmnist}
{\sc Yang, J., Shi, R., Wei, D., Liu, Z., Zhao, L., Ke, B., Pfister, H., and Ni, B.}
\newblock Medmnist v2-a large-scale lightweight benchmark for 2d and 3d biomedical image classification.
\newblock {\em Scientific Data 10}, 1 (2023), 41.

\bibitem{zeng2018natural}
{\sc Zeng, Z., Deng, Y., Li, X., Naumann, T., and Luo, Y.}
\newblock Natural language processing for ehr-based computational phenotyping.
\newblock {\em IEEE/ACM transactions on computational biology and bioinformatics 16}, 1 (2018), 139--153.

\bibitem{zhang2024curriculum}
{\sc Zhang, Z., Wang, J., and Zhao, L.}
\newblock Curriculum learning for graph neural networks: Which edges should we learn first.
\newblock {\em Advances in Neural Information Processing Systems 36\/} (2024).

\bibitem{zhu2004class}
{\sc Zhu, X., and Wu, X.}
\newblock Class noise vs. attribute noise: A quantitative study.
\newblock {\em Artificial intelligence review 22\/} (2004), 177--210.

\end{thebibliography}

\newpage
\begin{IEEEbiography}[{\includegraphics[width=1in,height=1.25in,clip,keepaspectratio]{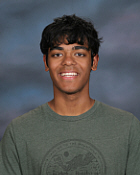}}]{Ruhaan Singh} is currently a research intern at the Oak Ridge National Laboratory, Oak Ridge, Tennessee and a student at Farragut High School, Knoxville, Tennessee. His research focuses on applying deep learning models to address pressing environmental problems. He is an International Science and Engineering Fair (ISEF) finalist, a USA Stockholm Junior Water Prize national finalist, and a two-time National Junior Science and Humanities Symposium (JSHS) finalist. He has presented his research at various international conferences including the American Geophysical Union (AGU), Neural Information Processing Systems (NeurIPS) and IEEE Big Data and has won the Best Machine Learning Innovation Award at NeurIPS and the AI Youth Star Award at IEEE Big Data.

\end{IEEEbiography}

\begin{IEEEbiography}[{\includegraphics[width=1.5in,height=2in,clip,keepaspectratio]{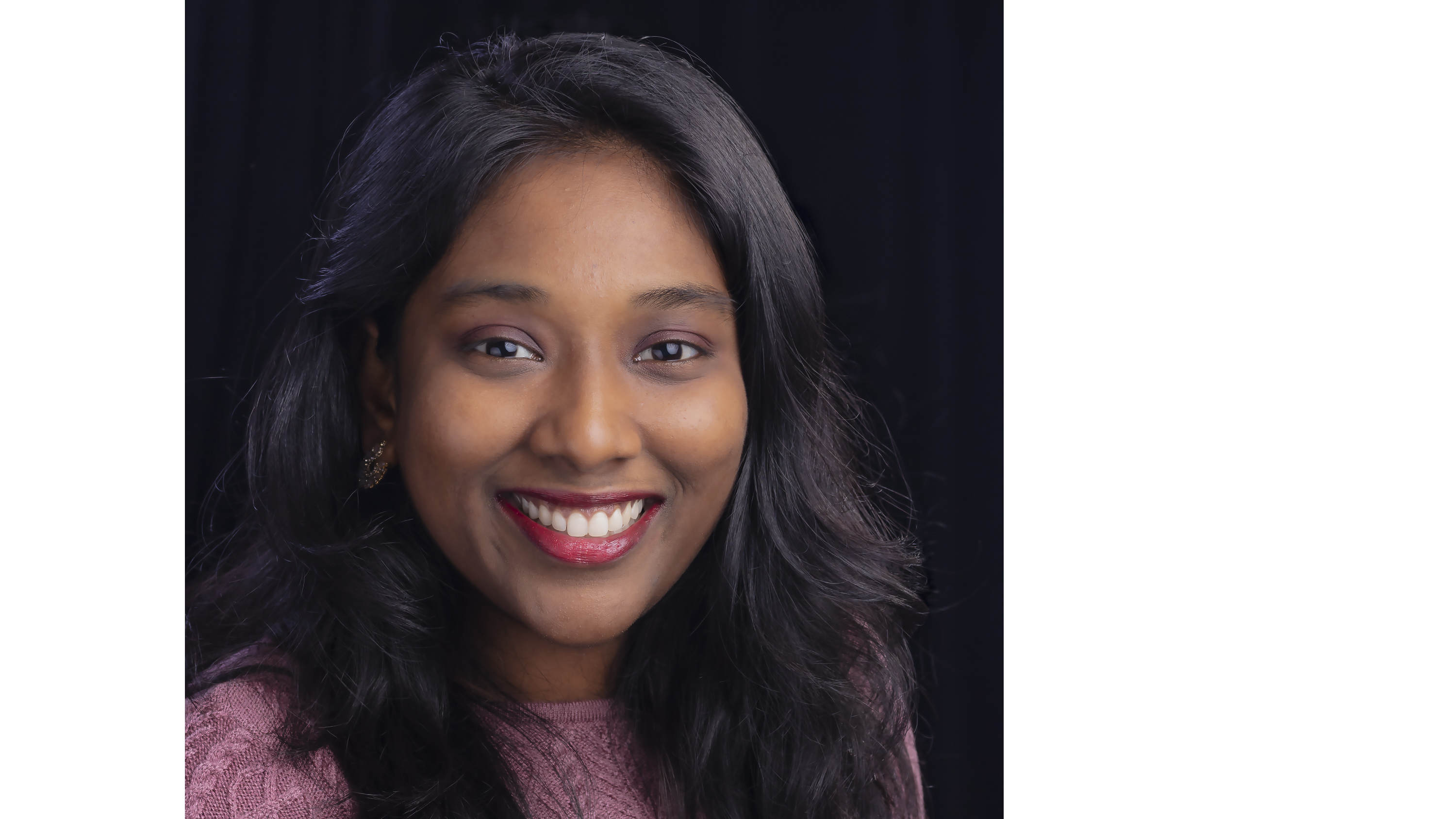}}]{Sreelekha Guggilam} (Member, IEEE) received the B.Math. degree (Hons.) in Mathematics from the Indian Statistical Institute, Bangalore, India, in 2014, the M.S. degree in Biostatistics-Bioinformatics from the University at Buffalo and Roswell Park Cancer Institute, Buffalo, NY, USA, in 2015, the M.S. degree in Civil Engineering (Transportation Statistics) from the University at Buffalo in 2017, and the Ph.D. degree in Computational Data Science from the University at Buffalo in 2022.

From 2022 to 2024, she was a Research and Development Associate in Machine Learning at Oak Ridge National Laboratory, Oak Ridge, TN, in the National Security Sciences Directorate, Geo-spatial Science and Human Dynamics Division.
Since August 2024, she has been an Assistant Professor in Data Science with the Department of Mathematics and Statistics at Texas A\&M University, Corpus Christi, TX, USA. 

Dr. Guggilam serves as a reviewer for multiple prestigious journals including Data Mining and Knowledge Discovery, Information Systems, Journal of Computational Science, Journal of Hydrology, IEEE Geoscience and Remote Sensing Letters, and Neural Networks. She is an active member of professional organizations including the Institute of Electrical and Electronics Engineers (IEEE), Association for Computing Machinery (ACM), American Geophysical Union (AGU), Society for Industrial and Applied Mathematics (SIAM), and Delta Omega Honorary Society in Public Health.
\end{IEEEbiography}
\EOD

\end{document}